\documentclass[12pt,english]{article}
\usepackage[T1]{fontenc}
\usepackage[latin9]{inputenc}
\usepackage[letterpaper]{geometry}
\usepackage{babel}

\usepackage[unicode=true, pdfusetitle,
 bookmarks=true,bookmarksnumbered=false,bookmarksopen=false,
 breaklinks=false,pdfborder={0 0 1},backref=false,colorlinks=false]
 {hyperref}

\makeatletter

\providecommand{\tabularnewline}{\\}

\makeatother

\begin{document}

\title{Evolving a New Feature for a Working Program}

\author{Mike Stimpson}
\maketitle
\begin{abstract}
\noindent A genetic programming system is created. A first fitness
function $f_{1}$ is used to evolve a program that implements a first
feature. Then the fitness function is switched to a second function
$f_{2}$, which is used to evolve a program that implements a second
feature while still maintaining the first feature. The median number
of generations $G_{1}$ and $G_{2}$ needed to evolve programs that
work as defined by $f_{1}$ and $f_{2}$ are measured. The behavior
of $G_{1}$ and $G_{2}$ are observed as the difficulty of the problem
is increased.

\noindent In these systems, the density $D_{1}$ of programs that
work (for fitness function $f_{1}$) is measured in the general population
of programs. The relationship $G_{1}\sim\frac{1}{\sqrt{D_{1}}}$ is
observed to approximately hold. Also, the density $D_{2}$ of programs
that work (for fitness function $f_{2}$) is measured in the general
population of programs. The relationship $G_{2}\sim\frac{1}{\sqrt{D_{2}}}$
is observed to approximately hold.
\end{abstract}

\section{INTRODUCTION}

Previous work \cite{Stimpson} demonstrated that, when evolving a
program from random starting programs, the relationship $G\sim\frac{1}{\sqrt{D}}$
often approximately held, where $G$ was the median number of generations
required to evolve a working program, and $D$ was the density of
working programs in the general population of programs. This paper
examines what happens when, after evolving a first feature, we attempt
to evolve a second feature of equivalent complexity.

The rest of this paper is organized as follows: Section 2 describes
the system. Section 3 presents the results in the form of several
data sets. Section 4 demonstrates the relationship between the density
of working programs and the median number of generations needed to
evolve a working program. Section 5 presents some conclusions, and
section 6 presents some open questions.

\section{\noindent THE SYSTEM: TREE-STRUCTURED PROGRAMS, SORTING INTEGERS}

I chose sorting a list of integers as the problem that programs were
attempting to solve. The first feature was sorting in ascending order;
the second feature was sorting in descending order.

The system had a fixed number $v$ of writable variables, numbered
1 through $v$. (\emph{Fixed} here means that it did not evolve; however,
it could be changed between runs via a command-line parameter.) It
also contained three read-only \textquotedbl{}variables\textquotedbl{}.
Variable 0 always contained 0. Variable $v+1$ always contained the
number of integers in the list being sorted. Variable $v+2$ contained
0 if the list was to be sorted in ascending order, and 1 otherwise.

A programs was represented as LISP-like tree structure. The trees
were limited to a maximum depth of 6. Programs contained a variable
number of nodes. Mutations could alter a whole sub-tree, rather than
a single node.

However, the programs were not purely in the LISP style, in that each
node could access any of the variables - variables were not sub-nodes
of operator nodes.

Statements were created from the following node types: \texttt{For},
\texttt{IfElse}, \texttt{CompareSwap}, and \texttt{ReverseCompareSwap}.
\texttt{CompareSwap} and \texttt{ReverseCompareSwap} were leaf nodes;
\texttt{For} and \texttt{IfElse} were not.

\texttt{For} was a C-style \texttt{for} loop with a loop variable,
a variable from which to initialize the loop variable, and a limit
variable to compare the loop variable to. It required one child node,
which it executed once for each iteration of the loop.

\texttt{IfElse} was an if/else on a variable. It required two child
nodes. If the variable was non-zero, it executed the \textquotedbl{}if\textquotedbl{}
node; otherwise, it executed the \textquotedbl{}else\textquotedbl{}
node. Either the \textquotedbl{}if\textquotedbl{} node or the \textquotedbl{}else\textquotedbl{}
node (or both) could be null (no operation).

\texttt{CompareSwap} compared two numbers in the list, and swapped
them if they were out of ascending order. \texttt{ReverseCompareSwap}
was identical, except that it swapped them if they were out of descending
order.

The difficulty was changed by increasing the number of variables,
which increased the odds of using the wrong variables when attempting
to create nested loops. That is, it decreased the probability of creating
a working program.

The population size was 1000 programs. Parents were chosen by a 7-way
tournament of randomly-chosen programs.

The fitness function $f_{1}$ was computed by having each program
attempt to sort three lists of numbers, which contained 10, 30 and
50 values. The lists contained the values from 1 to the size of the
list, in random order. After a program attempted to sort a list, the
\emph{forward distance} was computed as follows: For each location
in the list, the absolute value was taken of the difference between
the value at that location in the list as sorted by the program, and
the value that would be at that location if the list were perfectly
sorted in ascending order. A perfectly sorted list therefore had a
forward distance of zero. The \emph{reverse distance} was identical,
except that the perfectly sorted list was replaced by one that was
perfectly sorted in reverse (descending) order. In general, the forward
and backward distances were larger for the longer lists. To address
this, a \emph{normalized metric} was created for each list, which
was the reverse distance minus the forward distance, divided by the
sum of the forward and reverse distances. This evaluated to 1 for
a list perfectly sorted in ascending order, and to -1 for a list that
was perfectly sorted in descending order. Finally, $f_{1}$ was the
average of the normalized metrics for the three lists.

The fitness function $f_{2}$ ran the program twice, once with variable
$v+2$ cleared, and once with it set. For both runs, $f_{1}$ was
computed on the results. Call the results $f_{1a}$ and $f_{1d}$
for the runs that should sort in ascending and descending order, respectively.
Then, $f_{1d}=-1$ means that the program sorted perfectly in descending
order, and $f_{1d}=1$ means that the program failed as badly as possible
to sort in descending order. Then $f_{2}=\frac{f_{1a}-f_{1d}}{2}$
yields 1 if the lists were sorted correctly in both ascending and
descending order, and -1 if they were always sorted in the wrong order.

I also used the fitness function $f_{3}=\frac{(2f_{1a}-f_{1d})}{3}$.
This was similar to $f_{2}$, but it placed greater weight on preserving
the ability to sort in ascending order (that is, to preserve the functionality
that was already evolved by using $f_{1}$).

If the program executed 10 times as many statements as bubble sort
would require for the same list, the program was considered to be
in a semi-infinite loop, and terminated. No fitness penalty was imposed
for this condition.

The unsorted lists of numbers were randomly created. New lists were
created for each generation. The same lists were used for all programs
of any one generation.

After evolving a working program according to metric $f_{1}$, the
evolution was continued for ten more generations using $f_{1}$, in
order to reach something approaching a steady state, and yet not to
reach a monoculture. After these ten generations, approximately 96\%
of the programs had a perfect fitness function according to $f_{1}$.
Then the fitness function was switched to $f_{2}$ or $f_{3}$.

An \emph{evolution} started with a random collection of programs,
and proceeded until a program evolved that worked (had a fitness function
of 1.0). An evolution was characterized by the number of generations
required to evolve a working program as determined by $f_{1}$, and
the number of generations needed to evolve a working program according
to $f_{2}$ or $f_{3}$. The ten generations to approach steady state
were not included in these numbers. Also, the number of generations
for $f_{2}$ or $f_{3}$ did not include the generations when the
fitness function was $f_{1}$.

However, since evolution is a random process, a repeat of the evolution
would take a completely different number of generations.

A \emph{run} was 100 evolutions, all with the same parameters. It
was characterized by the median of the number of generations required
for the evolutions in the run. $G_{1}$ was the median number of generations
with fitness function $f_{1}$ (excluding the ten generations to approach
steady state); $G_{2}$ was the median number of generations with
fitness function $f_{2}$ or $f_{3}$. (The distribution of the number
of generations had a very long tail. The presence or absence of one
anomalous evolution could significantly shift the average, so the
median was the appropriate choice here.)

I also measured the density of working programs (as defined by $f_{1}$
or $f_{2}$ - note that $f_{3}$ gives the same definition of \textquotedbl{}working\textquotedbl{}
as $f_{2}$) in the general population of programs, by generating
a large number of random programs and seeing how many of them worked
as is, that is, with no evolution. I made sure that the sample was
large enough to contain at least 100 working programs.

The system presented a problem when measuring densities, because the
universe of all possible programs is not, in general, very much like
the set of programs that work. The universe of all possible programs
is weighted heavily toward the longest lengths, but the working programs
are not. As an evolution proceeds, the length distribution of the
population of programs should become more and more similar to the
distribution of working programs, and less and less similar to the
distribution of the universe of all possible programs. Given, then,
that the universe of all possible programs is structurally different
from both the working programs that are evolved and from the population
during an evolution, how can we get meaningful density data? I chose
the approach of trying to create self-consistent population distributions
- that is, population distributions such that, when populations with
that length distribution were evolved, the resulting working programs
had the same distribution of lengths. (In practice, this could only
be approximately achieved.) If we measure the density of a population
of programs with the same length distribution as the working programs,
we obtain density data that we can meaningfully combine with the median
number of generations. (The alternative - the density data coming
from populations that are unlike the population of working programs
- clearly is less likely to provide meaningful data.)

Finally, I measured the density of programs that worked as defined
by $f_{2}$ within the set of programs that worked as defined by $f_{1}$.
Again, I made sure that the sample was large enough to contain at
least 100 working programs (as defined by $f_{2}$).

\section{DATA AND ANALYSIS}

Generations to evolve a working program, using metric $f_{2}$:

\begin{tabular}{|c|c|c|}
\hline 
Number of variables & $G_{1}$ & $G_{2}$\tabularnewline
\hline 
2 & 1 & 62.5\tabularnewline
\hline 
3 & 4 & 82.5\tabularnewline
\hline 
4 & 5 & 175\tabularnewline
\hline 
5 & 7 & 129.5\tabularnewline
\hline 
6 & 16 & 212.5\tabularnewline
\hline 
7 & 21 & 239\tabularnewline
\hline 
8 & 38.5 & 458\tabularnewline
\hline 
9 & 51 & 720\tabularnewline
\hline 
10 & 78 & 462.5\tabularnewline
\hline
\end{tabular}

A second try with the same parameters:

\begin{tabular}{|c|c|c|}
\hline 
Number of variables & $G_{1}$ & $G_{2}$\tabularnewline
\hline 
2 & 1 & 72.5\tabularnewline
\hline 
3 & 3 & 92.5\tabularnewline
\hline 
4 & 5 & 84.5\tabularnewline
\hline 
5 & 7 & 228.5\tabularnewline
\hline 
6 & 11 & 272.5\tabularnewline
\hline 
7 & 30 & 349.5\tabularnewline
\hline 
8 & 35 & 346.5\tabularnewline
\hline 
9 & 53.5 & 337.5\tabularnewline
\hline 
10 & 57.5 & 463\tabularnewline
\hline
\end{tabular}

Generations to evolve a working program, using metric $f_{3}$:

\begin{tabular}{|c|c|c|}
\hline 
Number of variables & $G_{1}$ & $G_{2}$\tabularnewline
\hline 
2 & 1 & 60.5\tabularnewline
\hline 
3 & 2.5 & 107.5\tabularnewline
\hline 
4 & 5 & 255\tabularnewline
\hline 
5 & 6 & 203.5\tabularnewline
\hline 
6 & 16.5 & 410.5\tabularnewline
\hline 
7 & 21 & 470\tabularnewline
\hline 
8 & 43 & 613.5\tabularnewline
\hline 
9 & 67 & 718.5\tabularnewline
\hline 
10 & 104.5 & 863\tabularnewline
\hline
\end{tabular}

A second try with the same parameters:

\begin{tabular}{|c|c|c|}
\hline 
Number of variables & $G_{1}$ & $G_{2}$\tabularnewline
\hline 
2 & 1 & 69.5\tabularnewline
\hline 
3 & 1 & 105\tabularnewline
\hline 
4 & 4 & 152.5\tabularnewline
\hline 
5 & 6 & 292\tabularnewline
\hline 
6 & 11 & 196\tabularnewline
\hline 
7 & 27.5 & 457.5\tabularnewline
\hline 
8 & 28 & 379.5\tabularnewline
\hline 
9 & 46 & 859\tabularnewline
\hline 
10 & 74.5 & 794.5\tabularnewline
\hline
\end{tabular}

The different $G_{1}$ values between the data for $f_{2}$ and $f_{3}$
are statistical fluctuations. In all cases, $G_{1}$ was for programs
that were evolved using metric $f_{1}$. (Clearly the data contains
a lot of noise!)

$f_{3}$ took more generations than $f_{2}$ to evolve the same program.
This seems intuitively reasonable, since $f_{3}$ places a higher
value on preserving the existing functionality.

What happens if we don't use metric $f_{1}$ to evolve a solution
to a sub-problem? What if we just use metric $f_{2}$ or $f_{3}$
the whole way? Let us call the median number of generations $G_{2}^{'}$.

Generations to evolve a working program, using metric $f_{2}$ only:

\begin{tabular}{|c|c|}
\hline 
Number of variables & $G_{2}^{'}$\tabularnewline
\hline 
2 & 72.5\tabularnewline
\hline 
3 & 67\tabularnewline
\hline 
4 & 131.5\tabularnewline
\hline 
5 & 247\tabularnewline
\hline 
6 & 397\tabularnewline
\hline 
7 & 462.5\tabularnewline
\hline 
8 & 651.5\tabularnewline
\hline 
9 & 1003\tabularnewline
\hline 
10 & 1318.5\tabularnewline
\hline
\end{tabular}

\pagebreak{}Generations to evolve a working program, using metric
$f_{3}$ only:

\begin{tabular}{|c|c|}
\hline 
Number of variables & $G_{2}^{'}$\tabularnewline
\hline 
2 & 35.5\tabularnewline
\hline 
3 & 108\tabularnewline
\hline 
4 & 138\tabularnewline
\hline 
5 & 206\tabularnewline
\hline 
6 & 258\tabularnewline
\hline 
7 & 288.5\tabularnewline
\hline 
8 & 469\tabularnewline
\hline 
9 & 650.5\tabularnewline
\hline 
10 & 649\tabularnewline
\hline
\end{tabular}

Clearly, trying to evolve a working program using only $f_{2}$ took
more total generations than using $f_{1}$ and then $f_{2}$, but
using $f_{3}$ only took fewer total generations than using $f_{1}$
and then $f_{3}$. A possible reason for this is that $f_{2}$ is
symmetric - an initial random program is not likely to sort (even
partially) in both ascending and descending order, and a program that
(partially or completely) sorts only in ascending (or descending)
order gets a fitness of zero according to $f_{2}$. But $f_{3}$ gives
a positive value for a program that sorts (even partially) in ascending
order only. Programs can therefore begin evolving under $f_{3}$ more
easily than under $f_{2}$.

Density $D_{1}$ of fully-working programs (as measured by $f_{1}$)
in the general population of programs:

\begin{tabular}{|c|c|}
\hline 
Number of variables & $D_{1}$\tabularnewline
\hline 
2 & $1.107\times10^{-3}$\tabularnewline
\hline 
3 & $5.9\times10^{-4}$\tabularnewline
\hline 
4 & $3.1\times10^{-4}$\tabularnewline
\hline 
5 & $1.61\times10^{-4}$\tabularnewline
\hline 
6 & $1.16\times10^{-4}$\tabularnewline
\hline 
7 & $6.0\times10^{-5}$\tabularnewline
\hline 
8 & $4.3\times10^{-5}$\tabularnewline
\hline 
9 & $3.45\times10^{-5}$\tabularnewline
\hline 
10 & $2.06\times10^{-5}$\tabularnewline
\hline
\end{tabular}

\pagebreak{}Density $D_{2}$ of fully-working programs (as measured
by $f_{2}$ or $f_{3}$) in the general population of programs:

\begin{tabular}{|c|c|}
\hline 
Number of variables & $D_{2}$\tabularnewline
\hline 
2 & $4.4\times10^{-6}$\tabularnewline
\hline 
3 & $1.07\times10^{-6}$\tabularnewline
\hline 
4 & $3.3\times10^{-7}$\tabularnewline
\hline 
5 & $1.06\times10^{-7}$\tabularnewline
\hline 
6 & $5.37\times10^{-8}$\tabularnewline
\hline 
7 & $1.96\times10^{-8}$\tabularnewline
\hline 
8 & $1.06\times10^{-8}$\tabularnewline
\hline 
9 & $5.87\times10^{-9}$\tabularnewline
\hline 
10 & $2.47\times10^{-9}$\tabularnewline
\hline
\end{tabular}

But the evolution using metric $f_{2}$ was not done on a collection
of random programs; it was done on programs almost all of which were
fully working as defined by metric $f_{1}$. Perhaps, then, rather
than using $D_{2}$ (the density of programs that are fully working
under metric $f_{2}$ within the general population of programs),
we should use the density of programs that are fully working under
metric $f_{2}$ within the population of programs that are fully working
under metric $f_{1}$. Call this density $D_{2}^{'}$.

\begin{tabular}{|c|c|}
\hline 
Number of variables & $D_{2}^{'}$\tabularnewline
\hline 
2 & $4.31\times10^{-3}$\tabularnewline
\hline 
3 & $2.03\times10^{-3}$\tabularnewline
\hline 
4 & $1.111\times10^{-3}$\tabularnewline
\hline 
5 & $6.95\times10^{-4}$\tabularnewline
\hline 
6 & $4.68\times10^{-4}$\tabularnewline
\hline 
7 & $2.86\times10^{-4}$\tabularnewline
\hline 
8 & $2.21\times10^{-4}$\tabularnewline
\hline 
9 & $1.594\times10^{-4}$\tabularnewline
\hline 
10 & $1.244\times10^{-4}$\tabularnewline
\hline
\end{tabular}

\section{RELATIONSHIP BETWEEN SOLUTION DENSITY AND NUMBER OF GENERATIONS}

Combining the measured densities with the median number of generations
to reach a working program, we observe a pattern: As we change the
number of variables, the median number of generations needed to evolve
a working program is almost proportional to the reciprocal of the
square root of the density; that is, $K_{1}=G_{1}\times\sqrt{D_{1}}$
is almost constant. This value ($K_{1}$) rises slowly as $D_{1}$
decreases. But $K_{2}=G_{2}\times\sqrt{D_{2}}$ decreases slowly as
$D_{2}$ decreases.

\pagebreak{}Evolved using metric $f_{2}$:

\begin{tabular}{|c|c|c|c|c|c|c|}
\hline 
Number of variables & $G_{1}$ & $D_{1}$ & $K_{1}$ & $G_{2}$ & $D_{2}$ & $K_{2}$\tabularnewline
\hline 
2 & 1 & $1.107\times10^{-3}$ & 0.0333 & 62.5 & $4.4\times10^{-6}$ & 0.1311\tabularnewline
\hline 
3 & 4 & $5.9\times10^{-4}$ & 0.0972 & 82.5 & $1.07\times10^{-6}$ & 0.0853\tabularnewline
\hline 
4 & 5 & $3.1\times10^{-4}$ & 0.088 & 175 & $3.3\times10^{-7}$ & 0.1005\tabularnewline
\hline 
5 & 7 & $1.61\times10^{-4}$ & 0.0888 & 129.5 & $1.06\times10^{-7}$ & 0.0422\tabularnewline
\hline 
6 & 16 & $1.16\times10^{-4}$ & 0.1723 & 212.5 & $5.37\times10^{-8}$ & 0.0492\tabularnewline
\hline 
7 & 21 & $6.0\times10^{-5}$ & 0.1627 & 239 & $1.962\times10^{-8}$ & 0.0335\tabularnewline
\hline 
8 & 38.5 & $4.3\times10^{-5}$ & 0.252 & 458 & $1.06\times10^{-8}$ & 0.0472\tabularnewline
\hline 
9 & 51 & $3.45\times10^{-5}$ & 0.3 & 720 & $5.87\times10^{-9}$ & 0.0551\tabularnewline
\hline 
10 & 78 & $2.06\times10^{-5}$ & 0.354 & 462.5 & $2.47\times10^{-9}$ & 0.023\tabularnewline
\hline
\end{tabular}

A second try with the same parameters:

\begin{tabular}{|c|c|c|c|c|c|c|}
\hline 
Number of variables & $G_{1}$ & $D_{1}$ & $K_{1}$ & $G_{2}$ & $D_{2}$ & $K_{2}$\tabularnewline
\hline 
2 & 1 & $1.107\times10^{-3}$ & 0.0333 & 72.5 & $4.4\times10^{-6}$ & 0.1521\tabularnewline
\hline 
3 & 3 & $5.9\times10^{-4}$ & 0.0729 & 92.5 & $1.07\times10^{-6}$ & 0.0957\tabularnewline
\hline 
4 & 5 & $3.1\times10^{-4}$ & 0.088 & 84.5 & $3.3\times10^{-7}$ & 0.0485\tabularnewline
\hline 
5 & 7 & $1.61\times10^{-4}$ & 0.0888 & 228.5 & $1.06\times10^{-7}$ & 0.0744\tabularnewline
\hline 
6 & 11 & $1.16\times10^{-4}$ & 0.1184 & 272.5 & $5.37\times10^{-8}$ & 0.0631\tabularnewline
\hline 
7 & 30 & $6.0\times10^{-5}$ & 0.232 & 349.5 & $1.962\times10^{-8}$ & 0.049\tabularnewline
\hline 
8 & 35 & $4.3\times10^{-5}$ & 0.23 & 346.5 & $1.06\times10^{-8}$ & 0.0357\tabularnewline
\hline 
9 & 53.5 & $3.45\times10^{-5}$ & 0.314 & 337.5 & $5.87\times10^{-9}$ & 0.0259\tabularnewline
\hline 
10 & 57.5 & $2.06\times10^{-5}$ & 0.261 & 463 & $2.47\times10^{-9}$ & 0.023\tabularnewline
\hline
\end{tabular}

Evolved using metric $f_{3}$:

\begin{tabular}{|c|c|c|c|c|c|c|}
\hline 
Number of variables & $G_{1}$ & $D_{1}$ & $K_{1}$ & $G_{2}$ & $D_{2}$ & $K_{2}$\tabularnewline
\hline 
2 & 1 & $1.107\times10^{-3}$ & 0.0333 & 60.5 & $4.4\times10^{-6}$ & 0.1269\tabularnewline
\hline 
3 & 2.5 & $5.9\times10^{-4}$ & 0.0607 & 107.5 & $1.07\times10^{-6}$ & 0.1112\tabularnewline
\hline 
4 & 5 & $3.1\times10^{-4}$ & 0.088 & 255 & $3.3\times10^{-7}$ & 1465\tabularnewline
\hline 
5 & 6 & $1.61\times10^{-4}$ & 0.0761 & 203.5 & $1.06\times10^{-7}$ & 0.0663\tabularnewline
\hline 
6 & 16.5 & $1.16\times10^{-4}$ & 0.1777 & 410.5 & $5.37\times10^{-8}$ & 0.0951\tabularnewline
\hline 
7 & 21 & $6.0\times10^{-5}$ & 0.1627 & 470 & $1.962\times10^{-8}$ & 0.0658\tabularnewline
\hline 
8 & 43 & $4.3\times10^{-5}$ & 0.282 & 613.5 & $1.06\times10^{-8}$ & 0.0632\tabularnewline
\hline 
9 & 67 & $3.45\times10^{-5}$ & 0.394 & 718.5 & $5.87\times10^{-9}$ & 0.055\tabularnewline
\hline 
10 & 104.5 & $2.062\times10^{-5}$ & 0.475 & 863 & $2.47\times10^{-9}$ & 0.0429\tabularnewline
\hline
\end{tabular}

\pagebreak{}A second try with the same parameters:

\begin{tabular}{|c|c|c|c|c|c|c|}
\hline 
Number of variables & $G_{1}$ & $D_{1}$ & $K_{1}$ & $G_{2}$ & $D_{2}$ & $K_{2}$\tabularnewline
\hline 
2 & 1 & $1.107\times10^{-3}$ & 0.0333 & 69.5 & $4.4\times10^{-6}$ & 0.1458\tabularnewline
\hline 
3 & 1 & $5.9\times10^{-4}$ & 0.0243 & 105 & $1.07\times10^{-6}$ & 0.1086\tabularnewline
\hline 
4 & 4 & $3.1\times10^{-4}$ & 0.0704 & 152.5 & $3.3\times10^{-7}$ & 0.0876\tabularnewline
\hline 
5 & 6 & $1.61\times10^{-4}$ & 0.0761 & 292 & $1.06\times10^{-7}$ & 0.0951\tabularnewline
\hline 
6 & 11 & $1.16\times10^{-4}$ & 0.1185 & 196 & $5.37\times10^{-8}$ & 0.0454\tabularnewline
\hline 
7 & 27.5 & $6.0\times10^{-5}$ & 0.213 & 457.5 & $1.962\times10^{-8}$ & 0.0641\tabularnewline
\hline 
8 & 28 & $4.3\times10^{-5}$ & 0.1836 & 379.5 & $1.06\times10^{-8}$ & 0.0391\tabularnewline
\hline 
9 & 46 & $3.45\times10^{-5}$ & 0.27 & 859 & $5.87\times10^{-9}$ & 0.0658\tabularnewline
\hline 
10 & 74.5 & $2.062\times10^{-5}$ & 0.338 & 794.5 & $2.47\times10^{-9}$ & 0.0395\tabularnewline
\hline
\end{tabular}

But $K_{2}^{'}=G_{2}\times\sqrt{D_{2}^{'}}$ increases slowly as $D_{2}^{'}$
decreases.

Evolved using metric $f_{2}$:

\begin{tabular}{|c|c|c|c|}
\hline 
Number of variables & $G_{2}$ & $D_{2}^{'}$ & $K_{2}^{'}$\tabularnewline
\hline 
2 & 62.5 & $4.31\times10^{-3}$ & 4.1\tabularnewline
\hline 
3 & 82.5 & $2.03\times10^{-3}$ & 3.71\tabularnewline
\hline 
4 & 175 & $1.111\times10^{-3}$ & 5.83\tabularnewline
\hline 
5 & 129.5 & $6.95\times10^{-4}$ & 3.41\tabularnewline
\hline 
6 & 212.5 & $4.68\times10^{-4}$ & 4.6\tabularnewline
\hline 
7 & 239 & $2.86\times10^{-4}$ & 4.04\tabularnewline
\hline 
8 & 458 & $2.21\times10^{-4}$ & 6.81\tabularnewline
\hline 
9 & 720 & $1.594\times10^{-4}$ & 9.09\tabularnewline
\hline 
10 & 462.5 & $1.244\times10^{-4}$ & 5.16\tabularnewline
\hline
\end{tabular}

A second try with the same parameters:

\begin{tabular}{|c|c|c|c|}
\hline 
Number of variables & $G_{2}$ & $D_{2}^{'}$ & $K_{2}^{'}$\tabularnewline
\hline 
2 & 72.5 & $4.31\times10^{-3}$ & 4.76\tabularnewline
\hline 
3 & 92.5 & $2.03\times10^{-3}$ & 4.17\tabularnewline
\hline 
4 & 84.5 & $1.111\times10^{-3}$ & 2.82\tabularnewline
\hline 
5 & 228.5 & $6.95\times10^{-4}$ & 6.02\tabularnewline
\hline 
6 & 272.5 & $4.68\times10^{-4}$ & 5.9\tabularnewline
\hline 
7 & 349.5 & $2.86\times10^{-4}$ & 5.91\tabularnewline
\hline 
8 & 346.5 & $2.21\times10^{-4}$ & 5.15\tabularnewline
\hline 
9 & 337.5 & $1.594\times10^{-4}$ & 4.26\tabularnewline
\hline 
10 & 463 & $1.244\times10^{-4}$ & 5.16\tabularnewline
\hline
\end{tabular}

\pagebreak{}Evolved using metric $f_{3}$:

\begin{tabular}{|c|c|c|c|}
\hline 
Number of variables & $G_{2}$ & $D_{2}^{'}$ & $K_{2}^{'}$\tabularnewline
\hline 
2 & 60.5 & $4.31\times10^{-3}$ & 3.97\tabularnewline
\hline 
3 & 107.5 & $2.03\times10^{-3}$ & 4.84\tabularnewline
\hline 
4 & 255 & $1.111\times10^{-3}$ & 8.5\tabularnewline
\hline 
5 & 203.5 & $6.95\times10^{-4}$ & 5.36\tabularnewline
\hline 
6 & 410.5 & $4.68\times10^{-4}$ & 8.88\tabularnewline
\hline 
7 & 470 & $2.86\times10^{-4}$ & 7.94\tabularnewline
\hline 
8 & 613.5 & $2.21\times10^{-4}$ & 9.13\tabularnewline
\hline 
9 & 718.5 & $1.594\times10^{-4}$ & 9.07\tabularnewline
\hline 
10 & 863 & $1.244\times10^{-4}$ & 9.63\tabularnewline
\hline
\end{tabular}

A second try with the same parameters:

\begin{tabular}{|c|c|c|c|}
\hline 
Number of variables & $G_{2}$ & $D_{2}^{'}$ & $K_{2}^{'}$\tabularnewline
\hline 
2 & 69.5 & $4.31\times10^{-3}$ & 4.56\tabularnewline
\hline 
3 & 105 & $2.03\times10^{-3}$ & 4.73\tabularnewline
\hline 
4 & 152.5 & $1.111\times10^{-3}$ & 5.08\tabularnewline
\hline 
5 & 292 & $6.95\times10^{-4}$ & 7.7\tabularnewline
\hline 
6 & 196 & $4.68\times10^{-4}$ & 4.24\tabularnewline
\hline 
7 & 457.5 & $2.86\times10^{-4}$ & 7.73\tabularnewline
\hline 
8 & 379.5 & $2.21\times10^{-4}$ & 5.65\tabularnewline
\hline 
9 & 859 & $1.594\times10^{-4}$ & 10.8\tabularnewline
\hline 
10 & 794.5 & $1.244\times10^{-4}$ & 8.86\tabularnewline
\hline
\end{tabular}

\section{CONCLUSIONS}

Evolving the second feature (with metric $f_{2}$ or $f_{3}$) always
took more generations than the first feature (with metric $f_{1}$).
At best, it took 8 times as many generations. Evolving a new feature
into an already-working program is not easy; it is easier to evolve
the new feature as a separate program. That is, evolving sorting in
descending order is as easy as evolving sorting in ascending order.
But evolving sorting in descending order while preserving sorting
in ascending order is much harder. It's easier to evolve something
when you don't have to keep something else working.

$D_{2}^{'}\times D_{1}<D_{2}$ (slightly). That is, programs that
work according to $f_{2}$ are somewhat more abundant among programs
that work according to $f_{1}$ than one would expect merely from
knowing that all programs that work according to $f_{2}$ also work
according to $f_{1}$.

The relationship $G_{2}\sim\frac{1}{\sqrt{D_{2}}}$ approximately
holds when evolving a second feature within a population of programs
that implement a related first feature.

\section{FURTHER QUESTIONS}

What is the proportionality {}``constant''? (It's not really constant,
since it varies with population size, and maybe with other parameters.)

\end{document}